\documentclass[11pt]{article}

\usepackage[preprint]{acl}

\usepackage{times}
\usepackage{latexsym}
\usepackage[T1]{fontenc}
\usepackage[utf8]{inputenc}
\usepackage{microtype}
\usepackage{inconsolata}
\usepackage{graphicx}
\usepackage{amsmath}
\usepackage{amssymb}
\usepackage{bbm}
\usepackage{booktabs}
\usepackage{array}
\usepackage{booktabs}
\usepackage{tabularx}
\usepackage{enumitem}
\usepackage{adjustbox}
\usepackage{listings}
\newcommand{\ind}{\mathbbm{1}}

\title{Review Rarely Beats the Stronger Model:\\Separating Verification from Capability Transfer in LLM Pipelines}

\author{Yirui Zhang \qquad Yash Raj Shrestha \\
  University of Lausanne \\
  \texttt{\{yirui.zhang, yashraj.shrestha\}@unil.ch}}

\begin{document}
\maketitle

\begin{abstract}
Qwen3-8B reviewing Llama-3.1-8B in math adds 33.2
points, yet Qwen3-8B alone is 6.4 points more accurate than the result.  Reviewing another
model's answer is usually judged by its gain over the generator. We show that this
 confuses verification, which adds accuracy beyond either model alone, with capability transfer,
where a stronger reviewer supplies answers it could have given on its own. Measured against the stronger model alone, across 48 pairs of 12 open-weight models (3B
to 72B) in math, knowledge, code and reasoning, review beats it in only 9 of 96 review directions, and does worse the larger the capability gap. The reason is anchoring: the reviewer mostly follows the
generator, discarding its own correct answers. Review costs 2.6 to 2.8 times as much as the stronger model alone, while a selector using the same
two calls is 2.7 to 4.2 points more accurate than review  and cheaper. Studies of verification should report the stronger model alone, not only the gain over the generator.
\end{abstract}

\section{Introduction}
\label{sec:intro}
\begin{figure*}[t]
  \centering
  \includegraphics[width=0.92\textwidth]{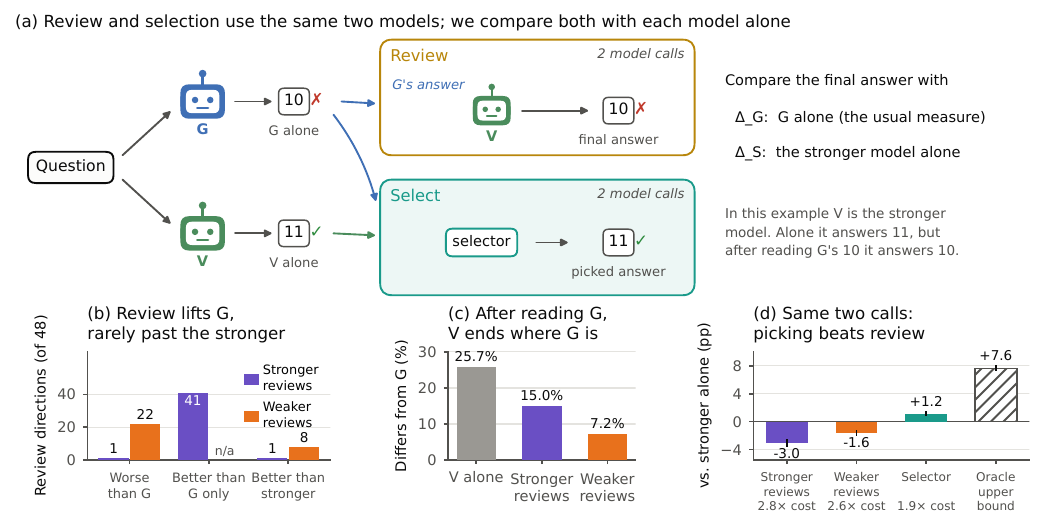}
  \caption{Overview. (a) In review, the reviewer (V) reads the generator's (G) answer and returns the final answer; in
  selection, a selector picks one of the two models' independent answers. Both use two calls. We compare the
  final answer with G alone ($\Delta_G$) and with the stronger model alone ($\Delta_S$).
  (b) Outcomes of the 48 directions of each kind. When the stronger model reviews, review beats G in 42 but the
  stronger model in only one; when the weaker one reviews (G is stronger), it beats both in 8 and is worse than G in 22.
  (c) After review, the final answer differs from G's in correctness far less often than V's own answer did.
  (d) With the same two calls, choosing between the two independent answers is more accurate than review; the oracle picks a correct answer when either model is correct.}
  \label{fig:overview}
\end{figure*}

\begin{figure}[t]
  \centering
  \includegraphics[width=\columnwidth]{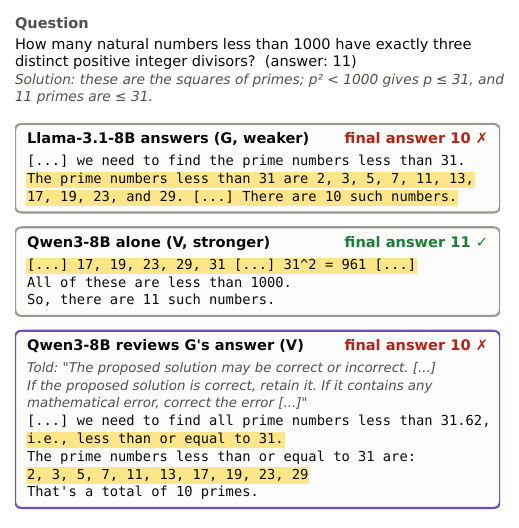}
  \caption{A reviewer that solves a question alone follows the generator's wrong answer after reading it.
  Llama-3.1-8B lists primes only up to 29 and answers 10. Qwen3-8B answers 11 (correct) on its own; reviewing, it
  writes the correct bound but copies the list that stops at 29, and answers 10.}
  \label{fig:example}
\end{figure}

As LLM outputs outpace human inspection \citep{lin2026outpace}, verification is increasingly delegated to models
themselves \citep{zheng2023judging}. Language model inference is costly in compute, money and energy
\citep{Luccioni2024power}, so a second model call is justified only if it buys accuracy that a single call cannot.
A common way to spend it is review, as when coding agents check each other's changes \citep{xiang2026crossmodel} or a multi-agent reasoning system adds a reviewer to its solver \citep{yang2026precise}.
This borrows a human practice: peer review, code review and second medical opinions all seek a second, independent
perspective, so placing a second model in the reviewer's seat is a natural design choice. A
generator (G) answers a question, and a reviewer (V) reads the question and G's full response and returns the final
answer, either keeping G's answer or correcting it \citep{welleck2023selfcorrect, xiang2026crossmodel}. Review, and
LLM verification more broadly, is usually judged by how much the final answer improves on G answering alone
\citep{welleck2023selfcorrect, lifshitz2025mav, zhou2026variation, lu2025verification}, and by that measure it often
looks very effective: in our math experiments, Qwen3-8B \citep{yang2025qwen3} reviewing Llama-3.1-8B
\citep{meta2024llama3} raises Llama-3.1-8B's accuracy by 33.2 points.

But review takes two model calls, and one of them already goes to the stronger model. When a stronger model reviews a weaker one, one would expect the result to be at least as accurate as the stronger model alone. Here the extra call buys less than nothing: Qwen3-8B answering alone is 6.4 points more accurate than the review's final answer. A gain over G can come from two very different sources: verification, which adds accuracy
beyond either model alone, and capability transfer, where a stronger reviewer supplies answers it could have given on
its own. The standard comparison cannot tell them apart and credits both to review. This hides a waste: a pipeline
may pay for review believing it verifies, while a single call to the stronger model would cost less and be at least as
accurate, and research may count capability transfer as progress in verification. In this paper, we separate the two
by measuring review in two ways (Figure~\ref{fig:overview}a): the gain over the generator ($\Delta_G$), the final
answer's accuracy minus G's accuracy alone, and the gain over the stronger model ($\Delta_S$), the final answer's
accuracy minus that of the more accurate of the two models answering alone.

A selector can only help where the second model is right and the stronger one wrong. We call
the share of
such questions the \textbf{headroom}: it is what an ideal choice between the two independent answers would add over the
stronger model. Review should be able to collect it, since the reviewer sees both the question and another model's
attempt.

We measure this on 48 pairs of 12 open-weight models from 3B to 72B parameters, with 2,500 math, 2,500 knowledge, 832
code and 901 reasoning questions, all graded by program. For every pair we run review in both directions, stronger reviewing weaker and weaker reviewing stronger, 96 review directions in total. We choose pairs so that both the gap,
the difference in accuracy between the two models answering alone, and the headroom vary widely. We ask what review
does (RQ1) and why (RQ2), whether a second model is worth its cost (RQ3), and whether to select or review (RQ4).

Our experiments show that review rarely beats the stronger model answering alone (Figure~\ref{fig:overview}b): only
9 of 96 directions do, seven of them a weaker model reviewing a close stronger one, and the larger the gap between the two
models, the worse review does. The reason is that the reviewer follows G: on questions that the reviewer answers
correctly alone and G gets wrong, the final answer is right only 40\% of the time when the stronger model reviews and
17\% when the weaker one does (Figure~\ref{fig:example}). After review, the final answer differs from G's in
correctness far less often than the two models alone do (Figure~\ref{fig:overview}c), so most of the 7.5 points of
headroom a pair leaves on average is lost. With the same two calls, a simple selector that picks one of the two
independent answers is 4.2 and 2.7 points more accurate than the two review directions and costs less in every pair
(Figure~\ref{fig:overview}d).

In summary, our key contributions are: 1) We reveal that judging review by its gain over the generator confuses
verification with capability transfer; 2) By splitting questions by which model is right alone, we show that review
falls short because the reviewer follows the generator's answer; 3) We propose an
evaluation framework with a second reference, a taxonomy of review outcomes and an equal-call baseline that any review
method should beat, and find that even a simple selector beats review at lower cost.

\section{Related Work}
\label{sec:related}
\begin{table*}[t]
  \centering
  \small
  \setlength{\tabcolsep}{3.5pt}
  \begin{tabularx}{\textwidth}{@{}l >{\raggedright\arraybackslash}X >{\raggedright\arraybackslash}p{1.45cm} >{\raggedright\arraybackslash}X
      >{\centering\arraybackslash}p{1.5cm} >{\centering\arraybackslash}p{1.25cm}
      >{\centering\arraybackslash}p{1.6cm}@{}}
    \toprule
    Work & Second model's role & Final answer & Reference (as defined in the paper) & Stronger model alone & Equal-call choice & Split by who is right alone \\
    \midrule
    \citet{huang2024cannot} & Same model revises its own answer & Same model & Accuracy after vs.\ before
      self-correction & n/a & \checkmark & n/a \\
    \citet{singhi2025solve} & Same model scores its own solutions & G's sample & Success rate vs.\ self-consistency
      at equal compute & n/a & \checkmark & n/a \\
    \citet{lifshitz2025mav} & Verifiers approve G's samples; most approved is kept & G's sample & Accuracy of
      BoN-MAV vs.\ pass@1 of G & when G is stronger & \checkmark & $\times$ \\
    \citet{lu2025verification} & Verifier accepts or rejects; G resamples & G's sample &
      $\mathrm{Precision}(S,V)-\mathrm{SolverAcc}(S)$ & $\times$ & $\times$ & $\times$ \\
    \citet{zhou2026variation} & Verifier filters G's samples & G's sample & $\Delta p_V = p_{G,V}-p_G$ & one case &
      $\times$ & $\times$ \\
    \citet{xiang2026crossmodel} & Other model reviews and rewrites & Reviewer & Pass rate vs.\ each model alone & one
      pair & $\times$ & $\times$ \\
    \midrule
    Ours & Other model reviews and rewrites & Reviewer & $\Delta_G$ and $\Delta_S$ (Section~\ref{sec:review}) &
      \checkmark & \checkmark & \checkmark \\
    \bottomrule
  \end{tabularx}
  \caption{Closest prior work. Most studies let the second model choose among the generator's own samples and measure
  the gain over the generator; in review, the second model writes the final answer. ``Equal-call choice'': a baseline
  that picks among independent answers at the same cost. ``Split by who is right alone'': outcomes split by whether
  each model answers correctly on its own.}
  \label{tab:related}
\end{table*}

\paragraph{Review and verification.}
We study zero-shot review by general-purpose models: a second model reads another model's full answer and returns the
final one. Trained correctors and verifiers \citep{welleck2023selfcorrect} are out of scope. A second model is a natural
choice, since without external feedback models struggle to correct their own reasoning \citep{kamoi2024selfcorrection}
and can even lose accuracy \citep{huang2024cannot}, and verifying a solution is often thought to be easier than
generating one \citep{zhou2026variation}. Table~\ref{tab:related} compares the closest studies. Most let the second
model approve or reject candidates sampled from the generator and measure the gain over the generator
\citep{lifshitz2025mav, lu2025verification, zhou2026variation}. In this form, several weak verifiers improve even
stronger generators \citep{lifshitz2025mav}: the generator is the stronger model, so this gain is also a gain over the
stronger model, and it comes from choosing among answers, in line with our finding that choosing succeeds where review
fails. \citet{xiang2026crossmodel} compare review with each model alone for a single pair on coding: Claude reviewing
Codex raises it from 71.6\% to 89.7\%, below Claude's own 91.4\%. None of these studies tests review against both the
stronger model alone and an equal-call choice, or traces where its gains come from. We do both across 96 review
directions in four domains: review's gains are mostly capability transfer.

\paragraph{Anchoring and conformity.}
Models are swayed by what they are shown: LLM judges by the order of candidates and by their own generations
\citep{wang2024fair, zheng2023judging, panickssery2024recognize}, and assistants by a user's stated views
\citep{sharma2023sycophancy}. Among peers, models follow a majority of scripted answers
\citep{zhu2025conformity, weng2025conformity}, and agreeing peers mislead a model that answered correctly far more
easily than they correct one that answered wrongly \citep{qu2026mislead}. In debates between real agents, agents defer
to a peer more than they keep their own answer \citep{choi2026identity}, and among models from 3B to 32B, smaller
models had more influence on the others' answer changes \citep{kasprova2026polite}. These peers are scripted answers,
copies of the same model or binary stances; in review, the peer is another model's full solution, across real
capability gaps in both directions. Review is also the minimal case: one model reads one other solution, with no
user, no majority, and a note that the solution may be incorrect, and anchoring persists even here. Prior work shows
that models change answers under pressure; we show what this costs a pipeline: most of the headroom a pair offers
is lost, which is why review loses to selection. Self-review shows that the cause is the other model's
answer, not the second pass: shown its own answer, a model keeps over 90\% of its right answers
(Appendix~\ref{app:checks}).

\paragraph{Selection, routing and equal cost.}
Methods that use more than one model differ in when they act. Routers decide which model should answer before any
answer exists \citep{ong2024routellm, ding2024hybrid, shnitzer2023routing}; cascades run a cheaper model first and
call a stronger one when its output looks unreliable \citep{chen2023frugalgpt, aggarwal2024automix, li2026jev};
ranking and fusion act after all candidates exist \citep{jiang2023blender}. Review and selection both act after the
answers exist: review shows one answer to the second model, while our selector, following selection-based ensembling
\citep{jiang2023blender}, picks one of the two independent answers. Comparisons at equal cost show that counting the
cost of verification changes the conclusion: multi-agent debate is no better than self-consistency at the same number
of responses \citep{huang2024cannot}, generative verification needs up to eight times the compute of self-consistency
to match it \citep{singhi2025solve}, and multi-agent verification overtakes both only with enough compute
\citep{lifshitz2025mav}. Commit-first judging has the judge write its own answer before using the candidate
\citep{zhou2026convincing, gozel2026commitfirst}; in one call, with the candidate already in the input,
we find that this answer is not independent of it (Appendix~\ref{app:checks}). The equal-cost comparisons above
choose among many samples of one generator. To our knowledge, no prior work uses choosing between two different
models' independent answers as the yardstick for review. Under it, review loses in both directions at the smallest
budget, two calls: selection is more accurate on average, never significantly less accurate in any pair, and cheaper
in every pair, with an external judge \citep{li2026jev} as a reference.

\section{Framework and Experimental Setup}
\label{sec:setup}

\subsection{Review Protocol and Comparative Gains}
\label{sec:review}

\paragraph{Generation and Review.}
In each review direction, every question is answered three times. The two models of a pair each answer alone with the same prompt, and in a
review one model (the generator, G) answers and the other (the reviewer, V) reads the question and G's full response
\citep{xiang2026crossmodel, yang2026precise}. V is told that the response may be right or wrong; it keeps the answer
if it judges it correct and otherwise corrects it, and either way returns a complete solution whose answer is the
final answer (GV; Appendix~\ref{app:prompts}).
Each pair is run in both review directions: the stronger model reviewing the weaker, and the weaker
reviewing the stronger. As sanity checks on a subset of questions, we also run a second
review prompt, which drops the instruction to keep a correct answer and has V commit to its own answer first, and self-review, in which V reviews its own solo answer instead of G's (Appendix~\ref{app:checks}).

\paragraph{Notation.}
For question $i$ of the $n$ questions of a direction, $g_i, v_i, f_i \in \{0,1\}$ indicate whether G alone, V alone
and the final answer are correct, and $s_i, w_i$ do the same for the stronger and the weaker model alone, ranked by
accuracy on the development part (Section~\ref{sec:tasks}). A bar is a mean over questions, so $\bar g$ is G's
accuracy alone.

\paragraph{Gain over the generator and the stronger model.}
\begin{equation}
  \Delta_G = \bar f - \bar g, \qquad \Delta_S = \bar f - \bar s .
\end{equation}
$\Delta_G$ is the number most studies report; $\Delta_S$ asks whether review beats simply calling the stronger
model. We call $\gamma = \bar s - \bar w$ the capability gap; it ranges from 0.5 to 39.6 points across our 48
pairs, and we report all differences in percentage points. Then $\Delta_G = \gamma + \Delta_S$ when the stronger
model reviews and $\Delta_G = \Delta_S$ when the weaker one does: a large $\Delta_G$ for a stronger reviewer can come entirely from the gap.

\paragraph{Gap, disagreement, and headroom.}
Two more quantities describe the two independent answers. The disagreement
$\delta = \frac{1}{n}\sum_i \ind[s_i \neq w_i]$ is the share of questions that exactly one model gets right. The
headroom
\begin{equation}
  h = \tfrac{1}{n}\textstyle\sum_i \ind[w_i=1,\,s_i=0]
\end{equation}
is what always picking a correct one of the two answers would add over the stronger model. Since
$\delta = \gamma + 2h$, $h = (\delta - \gamma)/2$.

\paragraph{Decomposing review gains by question type.}
Grouping the questions by $(g_i, v_i)$ splits the gain over the generator into three parts:
\begin{equation}
\begin{aligned}
\Delta_G
={}& \underbrace{\tfrac{1}{n}\textstyle\sum\limits_{g_i=0,\,v_i=1} f_i}_{\text{V-only correct}}
 - \underbrace{\tfrac{1}{n}\textstyle\sum\limits_{g_i=1,\,v_i=0} (1-f_i)}_{\text{G-only correct}} \\
 &+ \underbrace{\tfrac{1}{n}\textstyle\sum\limits_{g_i=v_i}(f_i-g_i)}_{\text{same outcome}} .
\end{aligned}
\end{equation}
The parts are the reviewer's correct answers that the final answer takes up, G's correct answers that review
loses, and what review changes where the two models agree. The first part is capability transfer; only a positive
$\Delta_S$ shows verification.

\paragraph{Taxonomy of Review Outcomes.}
From the paired 95\% intervals $I_G$ and $I_S$ of $\Delta_G$ and $\Delta_S$ (Section~\ref{sec:stats}), we place
each direction at one of five outcome levels: worse than G ($I_G<0$), no different from G ($0\in I_G$), above G
but below the stronger model ($I_G>0$, $I_S<0$), level with the stronger model ($I_G>0$, $0\in I_S$), and beyond it
($I_S>0$, which takes precedence).

\subsection{Tasks and Models}
\label{sec:tasks}

All four domains are graded by program, never by an LLM judge (Table~\ref{tab:tasks}). In math and knowledge, 1,000
random questions form the development part used to rank models; other domains use all questions. An
unparseable answer counts as wrong; if the reviewer's answer cannot be parsed, or in code contains no program, the
final answer is G's.

\begin{table}[t]
  \centering
  \small
  \setlength{\tabcolsep}{3pt}
  \begin{tabular}{@{}llrl@{}}
    \toprule
    Domain & Source & $N$ & Grading \\
    \midrule
    Math & MATH, levels 3--5 & 2,500 & symbolic equiv. \\
    Knowledge & MMLU-Pro & 2,500 & option letter \\
    Code & BigCodeBench-Instruct & 832 & unit tests \\
    Reasoning & ReClor, FOLIO v2, GPQA & 901 & option letter \\
    \bottomrule
  \end{tabular}
  \caption{Tasks, from MATH \citep{hendrycks2021math}, MMLU-Pro \citep{wang2024mmlupro}, BigCodeBench
  \citep{zhuo2025bigcodebench}, and for reasoning 500 ReClor \citep{yu2020reclor}, 203 FOLIO v2 \citep{han2024folio}
  and 198 GPQA Diamond \citep{rein2024gpqa} questions.}
  \label{tab:tasks}
\end{table}

\paragraph{Models and pairs.}
We use 12 open-weight instruction-tuned models from 3B to 72B parameters: Qwen3 (4B, 8B, 32B; \citealp{yang2025qwen3}),
Qwen2.5 (3B, 7B, 72B; \citealp{yang2024qwen25}), Qwen2.5-Coder-7B \citep{hui2024qwen25coder},
Seed-Coder-8B \citep{seed2025seedcoder}, Llama-3.2-3B \citep{meta2024llama32}, Llama-3.1-8B \citep{meta2024llama3},
Ministral-8B \citep{mistral2024ministral} and Mistral-Small-24B \citep{mistral2025small3}.
Within each domain, we sort the models into strong, middle and weak tiers by development accuracy and, for each of
the six tier combinations, take one pair with high and one with low headroom, so that gap and headroom both vary
widely (Appendix~\ref{app:pairs}). This gives 12 pairs, or 24 review directions, per domain.

All models decode greedily, with thinking off for Qwen3 and up to 4,096 output tokens, so each model has one fixed
answer per question. Only 1.7--1.8\% of review outputs and 1.8--3.3\% of solo answers reach this limit
(Appendix~\ref{app:checks}). As a sanity check, we rerun Qwen3-8B with thinking on in two math directions
(Appendix~\ref{app:checks}). Prompts are in Appendix~\ref{app:prompts}.

\subsection{Statistical Inference and Uncertainty}
\label{sec:stats}

Our 95\% intervals resample questions with replacement, keeping each question's paired outcomes as one unit, so they
measure the uncertainty from which questions were sampled; averages weight the four domains equally. To relate
outcomes to the gap, we regress $\Delta_S$ and $\Delta_G$ on $\gamma$, alone and jointly with $h$, with one intercept per domain, resampling
pairs as well, and read the slope as a linear association across the evaluated pairs (details in
Appendix~\ref{app:robustness}).

\section{Results}
\label{sec:results}

\subsection{RQ1: What does review do?}
\label{sec:rq1}

\begin{figure*}[t]
  \centering
  \includegraphics[width=0.88\textwidth]{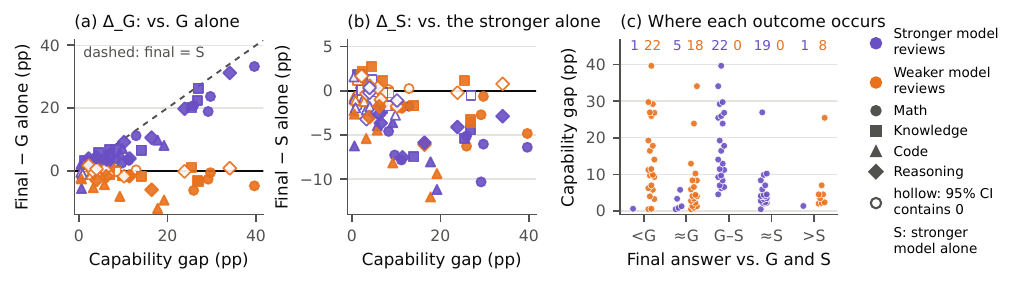}
  \caption{Each point is one review direction, coloured by who reviews whom; shape gives the domain, and hollow
  markers have a 95\% interval containing zero. (a) $\Delta_G$ against the gap; on the dashed line the final answer
  matches the stronger model alone. (b) $\Delta_S$ against the gap. (c) The gap of every direction at each outcome
  level; numbers count the directions of each colour.}
  \label{fig:gap}
\end{figure*}

By the paired 95\% intervals, the 96 review directions fall into the five levels as follows: worse than G (23),
no different from G (23), above G but below the stronger model (22), level with the stronger model (19), and beyond it
(9). Which level a direction reaches is largely decided by who reviews whom and by the gap (Figure~\ref{fig:gap}c; by gap range,
Figure~\ref{fig:levels}).

\paragraph{A stronger reviewer lifts G toward itself, not past it.}
Measured against the generator (Figure~\ref{fig:gap}a), review looks very effective: the gain grows by 8.2 points
per 10 points of gap [7.0, 8.9] and reaches 33.2 points when Qwen3-8B reviews Llama-3.1-8B in math. Measured against
the stronger model (Figure~\ref{fig:gap}b), the same directions fall below zero, on average 3.1 points [2.7, 3.5]
below the reviewer answering alone, and 6.4 points for that math pair. The larger the gap, the more review lifts G and the further the result
falls below the stronger model: a large gain over G mostly measures what the reviewer could already do.

\begin{figure*}[!t]
  \centering
  \includegraphics[width=\textwidth]{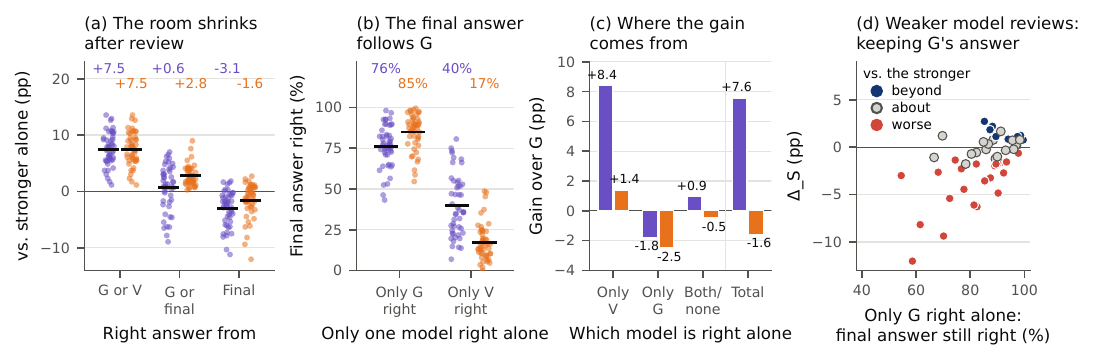}
  \caption{Why review leaves the second model's value unused. Each dot is one review direction, coloured as in Figure~\ref{fig:gap}; black bars are means with domains weighted equally. (a) Accuracy relative to the stronger model alone of the oracle of the two solo
  answers (G or V right), the oracle after review (G's answer or the final answer right), and the final answer.
  (b) On questions only one model gets right alone, how often the final answer is right. (c) The change over G alone,
  by which model is right alone. (d) Weaker model reviews: on questions only G gets right alone, how often the final
  answer stays right, against $\Delta_S$; colours give the outcome level.}
  \label{fig:mechanism}
\end{figure*}

\paragraph{A weaker reviewer often makes G worse.}
Here G is the better model, so $\Delta_G = \Delta_S$. Twenty-two of the 48 directions end significantly below G alone,
and the average change is $-1.6$ points [$-1.8$, $-1.3$]. Damage can happen at any gap but becomes the rule as the gap
grows: 10 of the 13 directions with a gap of 14 points or more are worse, against 12 of the 35 closer ones
(Figure~\ref{fig:gap}c). In code, Llama-3.2-3B reviewing Qwen3-8B lowers Qwen3-8B's accuracy by 12.0 points.

\paragraph{Beating the stronger model is rare.}
Only 9 of 96 directions beat the stronger model alone, all in math and knowledge. Eight are a weaker model reviewing
a stronger one, seven of them with a gap of at most 7 points. Being close makes a gain possible but does not deliver it: of the 26
weaker-reviews-stronger directions with a gap under 8 points, 7 beat the stronger model and 7 are significantly
worse.

\paragraph{Outcomes track the gap, not the headroom.}
Across the evaluated pairs, each 10 points of gap goes with a $\Delta_S$ lower by 1.8 points [1.1, 3.0] when the
stronger model reviews and by 1.5 points [0.7, 2.5] when the weaker one does. One might expect the headroom to matter
instead, but it is tied to the gap by construction ($h = (\delta - \gamma)/2$) and strongly negatively correlated with
it within each domain ($r = -0.73$ [$-0.84$, $-0.58$]): close pairs leave the most room. With the gap held fixed,
headroom has no detectable effect ($-0.8$ per 10 points when the stronger model reviews and $+2.3$ when the weaker one
does, both intervals containing zero), while the gap keeps its slope when the stronger model reviews (2.0
[0.8, 3.4]) and most of it when the weaker one does (1.0 [0.0, 2.4], an interval that just reaches zero). Close pairs
have the most to gain from a second opinion, yet even there review usually fails to collect it.

\subsection{RQ2: Why does review miss the gains a second model could offer?}
\label{sec:rq2}

\paragraph{Review removes the room a second model leaves.}
In all 48 pairs, the share of questions that at least one model answers correctly alone (the oracle) exceeds the
stronger model's accuracy, by 7.5 points on average [7.2, 7.8]. Review removes most of this room
(Figure~\ref{fig:mechanism}a): even picking, question by question, between G's own answer and the final answer would
add only 0.6 points [0.2, 1.0] over the stronger model when it reviews, and 2.8 [2.6, 2.9] when the weaker one does.
In 16 of the 48 stronger-reviews-weaker directions, this oracle even falls below the stronger model alone.

\paragraph{The final answer stays with G.}
On questions where exactly one model is right alone (Figure~\ref{fig:mechanism}b), if G was the one, the final answer
stays right 76\% of the time when the stronger model reviews and 85\% when the weaker one does. If V was the one, the
final answer is right only 40\% and 17\% of the time: a reviewer that solves a question alone often fails to once it
has read a wrong answer, even when it is the stronger model (Figure~\ref{fig:example}). Most of these lost answers
become G's wrong one (53\% and 65\%). Which ones it loses depends most on the gap: the further the reviewer is above G,
the more it keeps, and the further below, the fewer. Within a direction, it loses more when its own solution is long
and its maximum logit low, but not when G's answer is longer or more confident (Appendix~\ref{app:checks}). A second pass does not
explain this: shown its own answer in place of G's, a reviewer keeps over 90\% of these right answers. Nor does the wording of our prompt: with another prompt that has the reviewer commit first, the reviewer still kept only 55\%, against 61\% with ours. Shown its own answer next to G's in a third call, it recovers many of them but also goes back to many of its own wrong answers, and ends below the selector of Section~\ref{sec:cost} (Appendix~\ref{app:checks}). With thinking on, in two math directions, review still ends about 3 points below the reviewer alone, and appears to follow a stronger G but not a weaker one (Appendix~\ref{app:checks}).
Over all
questions, the two models alone differ in correctness on 25.7\% of them on average; the final answer
differs from G's on only 15.0\% when the stronger model reviews and 7.2\% when the weaker one does, with a drop in all
96 directions (Figure~\ref{fig:overview}c; per direction in
Figure~\ref{fig:convergence}, Appendix~\ref{app:figures}). Headroom lives only where the two models
differ, so once the final answer converges to G's, it is mostly lost.

\begin{figure*}[!t]
  \centering
  \includegraphics[width=0.861\textwidth]{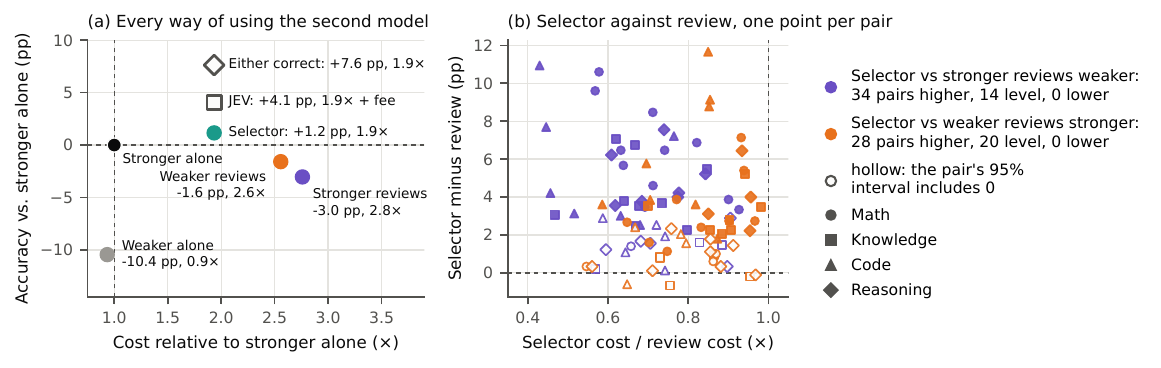}
  \caption{The cost and accuracy of using a second model; cost per question is parameters $\times$ (input + output
  tokens). (a) Each way of using the two models relative to the stronger model alone: means over 48 pairs with domains
  weighted equally, with 95\% intervals; JEV also pays an API fee not on this axis. (b) The selector against each
  review direction, one point per pair; hollow points have a 95\% interval that includes zero.}
  \label{fig:cost}
\end{figure*}

\paragraph{Review itself creates little.}
Where the two models agree, review can only fix a question both got wrong or break one both got right, and the net
is small: $+0.9$ points [$+0.7$, $+1.2$] when the stronger model reviews and $-0.5$ [$-0.7$, $-0.3$] when the weaker
one does. Splitting the questions into eight cells by whether G alone, V alone and the final answer are right (Figure~\ref{fig:mechanism}c, Table~\ref{tab:eight-cells}), a stronger reviewer's 7.6-point gain is
$+8.4$ on questions only V could solve, $-1.8$ on those only G could solve and $+0.9$ on the rest: almost all of it is capability transfer, not verification. For a weaker reviewer the parts are $+1.4$, $-2.5$ and $-0.5$.

\paragraph{Between close models, keeping G's right answers decides.}
When a weaker model reviews a close stronger one, the directions that beat the stronger model and those that hurt it
do not differ in headroom (8.7 vs.\ 7.4 points) or in how often the final answer takes up the reviewer's right
answers (23\% vs.\ 25\%). They differ in what happens to G's own right answers: on questions only G gets right
alone, the final answer is still right 92\% of the time in the first group and 72\% in the second, and across all 48
weaker-reviews-stronger directions this rate tracks $\Delta_S$ closely (Figure~\ref{fig:mechanism}d). Beating the
stronger model comes from not breaking anything, not from collecting headroom.

\paragraph{The gap sets how much is at stake.}
The gap sets how many questions only the stronger model gets right alone: 30\% of all questions when the two models
are 14 or more points apart, against 14\% when they are closer. These are the questions on which review can gain or
lose. A stronger reviewer carries its right answer into the final answer on 60\% of them in far-apart pairs, against
33\% in closer ones, so G gains more, yet stays below the reviewer alone. A weaker reviewer keeps G's right answer
about as often in both (87\% and 84\%), so it is no more persuasive in far-apart pairs; the same loss rate simply
hits twice as many questions.

\section{Using the second model}
\label{sec:cost}

\paragraph{Cost.}
We count the cost of a model call with $N$ parameters that reads $t_{\text{in}}$ and writes $t_{\text{out}}$ tokens as
\begin{equation}
  c = N \, (t_{\text{in}} + t_{\text{out}}).
\end{equation}
This is proportional to forward-pass compute and puts a 3B and a 72B model on one scale. A review costs $c_G + c_V$,
with G's whole answer in V's input; the selector costs the two solo calls. Appendix~\ref{app:cost} repeats every
comparison with $c' = N\,(t_{\text{in}} + 4\,t_{\text{out}})$.

\paragraph{Selector.}
The selector lets both models answer alone and picks one answer with no further model call, so it uses the same two
calls as review. A small classifier, with features and settings fixed in advance and fitted separately for each pair, estimates from numbers
alone (token log-probabilities, lengths, whether the two answers agree; Appendix~\ref{app:selector}) the probability
that each answer is right, and the selector returns the weaker model's answer only if its estimate is higher. It is
always evaluated on questions it was not fitted on, so its math and knowledge numbers use 1,500
held-out questions and differ slightly from Section~\ref{sec:results}. Its labels come from the same development
questions that are needed anyway to know which model is stronger; a rule with no labels, returning the answer with
the higher mean token log-probability, is 3.7 points below the stronger model alone (``A selector without
labels'', Appendix~\ref{app:checks}).

\paragraph{A stronger reference.}
An external LLM judge, JEV \citep{li2026jev}, reads the question and both full answers and estimates
which one is right (Appendix~\ref{app:selector}). It is a closed API priced in dollars, so we report it beside the
selector, not against review.

\subsection{RQ3: Is a second model worth its cost?}
\label{sec:rq3}

\paragraph{Review costs more and ends below the stronger model; selection ends above it.}
Figure~\ref{fig:cost}a places each way by its cost and accuracy relative to the stronger
model alone. Review costs 2.8 times as much for $-3.0$ points [$-3.5$, $-2.6$] when the stronger model reviews, and
2.6 times for $-1.6$ [$-1.9$, $-1.3$] when the weaker one does; it is significantly above the stronger model in 1 and
7 of the 48 pairs and below it in 23 and 21. The selector costs 1.9 times as much and gains 1.2 points [0.9, 1.4],
significantly above the stronger model in 19 pairs and below it in none, recovering about 15\% of the 7.6 points
[7.3, 7.9] of room. JEV recovers more, 4.1 points [3.8, 4.3], above the stronger model in 42 of 48 pairs, for an API fee of \$0.04--\$0.21 per 1,000 questions.

Review is expensive because the reviewer reads G's whole answer, on average longer than its entire solo
call, and still writes 0.8 to 1.0 times as much as alone. By
domain (Table~\ref{tab:domain-cost}), the selector is above the stronger model everywhere except code ($+0.5$
[$-0.1$, $+1.0$]); review is above it in no domain.

\subsection{RQ4: Select or review?}
\label{sec:rq4}

\paragraph{At the same two calls, choosing beats review.}
Pair by pair (Figure~\ref{fig:cost}b), the selector is 4.2 points [3.8, 4.6] more accurate than the stronger model
reviewing the weaker, higher in 34 of 48 pairs and lower in none, at 0.69 times the cost. Against the weaker model
reviewing the stronger, it is 2.7 points [2.4, 3.1] more accurate, higher in 28 pairs and lower in none, at 0.82 times
the cost. It is cheaper in every pair and more accurate in every domain. Weighting output tokens four times brings the
costs close (0.89 and 0.98 times).

\section{Conclusion}
\label{sec:discussion}

Gain over the generator mixes verification with capability transfer. Once the two are separated, 96 review
directions in four domains tell one story: review rarely beats the stronger model alone, because a reviewer that
reads the generator's answer tends to follow it, so a second model is better used for its own answer, as a
candidate to choose from, than as a reviewer. For researchers, this means reporting the stronger model alone
and the decomposition by which model is right alone (Section~\ref{sec:review}) alongside $\Delta_G$, since
$\Delta_G$ on its own credits a stronger reviewer's own ability to review. For practitioners, it means checking,
before adding a review step, whether a single call to the better model already wins; if a second call is worth
paying for, choosing between the two models' answers is the baseline to beat. LLM review inherits the form of
human review but not reliably the benefit of an independent view: the reviewer often has the right answer on its
own, yet gives it up after reading G's, even when asked to commit to its own answer first.

\section*{Limitations}

In this paper, we showed that the standard evaluation of review confuses verification with capability transfer, a
distinction that is essential for deciding whether a second model call is worth its cost. Some limits of scope remain.

\paragraph{Other models.} We study open-weight models from 3B to 72B parameters, and Qwen3 models with thinking on in a few math
directions. Testing closed frontier models and more thinking-on models is
left for future work.

\paragraph{Other tasks.} All our tasks are in English and graded by program. Extending the comparison to
open-ended generation, where no program can grade an answer, is left for future research.

\paragraph{Why the reviewer follows G.} We measured how often the final answer follows G and what this costs a
pair, checked one alternative prompt on four directions (in math, the reviewer often skipped committing to its own
answer first) and a third call that shows the reviewer its own answer next to G's, but not what in the reviewer
causes it. Other prompts, longer exchanges between the two models and training are left for future work. Likewise, verification and capability transfer are defined by
outcome, not by what happens inside the reviewer: a right answer the reviewer also gives alone counts as transfer,
even if it was reached by checking G's.

\bibliography{custom}

\appendix

Code and per-question results are available at \url{https://github.com/Kamilaabb/verification-or-capability-transfer}.
Models are served with vLLM 0.29.0 \citep{kwon2023vllm}; math answers are graded with Math-Verify 0.9.0
\citep{mathverify}; selectors use scikit-learn 1.9.1 \citep{pedregosa2011sklearn}; code is run in the official
BigCodeBench evaluation environment \citep{zhuo2025bigcodebench}.

\section{Model pairs}
\label{app:pairs}
Within each domain, we sort the models by their accuracy on the development questions (for code and reasoning, all
questions) into strong, middle and weak tiers. For a pair, we count the development questions that only the weaker
model answers correctly, which is the pair's headroom on those questions, and in each of the six tier combinations
we take the pair with the most and the pair with the fewest. Table~\ref{tab:pairs} lists the 48 pairs.

A few pairs needed a further rule, fixed before any test-set review results; Table~\ref{tab:pair-exceptions}
lists them. Where the chosen pair included Qwen2.5-72B, we took the next pair in headroom order to save compute, as
reviewing with it takes two H100 GPUs. We later ran the four original pairs in both directions as well;
Appendix~\ref{app:robustness} includes them in the analysis over all completed pairs.

\begin{table}[!ht]
  \centering
  \small
  \begin{tabular}{@{}>{\raggedright\arraybackslash}p{0.26\linewidth}>{\raggedright\arraybackslash}p{0.30\linewidth}>{\raggedright\arraybackslash}p{0.32\linewidth}@{}}
    \toprule
    Cell & Case & Pair used \\
    \midrule
    \multicolumn{3}{@{}l}{\emph{Ties and small tiers}} \\
    Knowledge, strong-middle, most & Two pairs tie & The pair already run on the development questions \\
    Knowledge, strong-weak, fewest & Two pairs tie & The pair whose weaker model is weaker \\
    Reasoning, middle-middle & The middle tier has two models & Its one pair \\
    Reasoning, strong-middle, fewest & Two pairs tie & Both (12 pairs in all) \\
    Knowledge, strong-strong, fewest & Only three strong models; without Qwen2.5-72B one pair is left
      & Qwen2.5-72B with Mistral-Small-24B, kept \\
    \midrule
    \multicolumn{3}{@{}l}{\emph{Qwen2.5-72B replaced by the next pair}} \\
    Math, strong-strong, most & Qwen2.5-72B with Qwen3-4B (79) & Qwen3-4B with Qwen3-8B (70) \\
    Math, strong-strong, fewest & Qwen2.5-72B with Qwen3-32B (35) & Qwen3-8B with Qwen3-32B (39) \\
    Knowledge, strong-middle, fewest & Qwen2.5-72B with Qwen2.5-7B (48) & Mistral-Small-24B with Qwen2.5-7B (64) \\
    Knowledge, strong-weak, fewest & Qwen2.5-72B with Llama-3.2-3B (35) & Mistral-Small-24B with Llama-3.2-3B (44) \\
    \bottomrule
  \end{tabular}
  \caption{Pairs chosen by a further rule. Numbers in parentheses: development questions that only the weaker model
  answers correctly.}
  \label{tab:pair-exceptions}
\end{table}

\begin{table*}[t]
  \centering
  \small
  \begin{tabular}{@{}lllllllrr@{}}
    \toprule
    Domain & Tiers & Headroom & Stronger & Acc. & Weaker & Acc. & Gap & $h$ \\
    \midrule
Math & strong-strong & most & Qwen3-8B & 80.3 & Qwen3-4B & 78.3 & 2.0 & 6.3 \\
 & strong-strong & fewest & Qwen3-32B & 85.1 & Qwen3-8B & 80.6 & 4.5 & 4.2 \\
 & strong-middle & most & Qwen3-4B & 78.3 & Mistral-Small-24B & 65.4 & 12.9 & 5.4 \\
 & strong-middle & fewest & Qwen3-32B & 85.1 & Qwen2.5-3B & 59.2 & 25.9 & 1.7 \\
 & strong-weak & most & Qwen3-8B & 80.6 & Ministral-8B & 50.8 & 29.7 & 2.3 \\
 & strong-weak & fewest & Qwen3-8B & 80.6 & Llama-3.1-8B & 40.9 & 39.6 & 1.2 \\
 & middle-middle & most & Qwen2.5-7B & 68.8 & Mistral-Small-24B & 65.4 & 3.5 & 8.5 \\
 & middle-middle & fewest & Qwen2.5-7B & 68.8 & Qwen2.5-3B & 59.2 & 9.7 & 6.1 \\
 & middle-weak & most & Qwen2.5-3B & 59.2 & Ministral-8B & 50.8 & 8.3 & 7.9 \\
 & middle-weak & fewest & Qwen2.5-7B & 68.8 & Llama-3.2-3B & 39.6 & 29.2 & 3.0 \\
 & weak-weak & most & Llama-3.1-8B & 40.9 & Llama-3.2-3B & 39.6 & 1.3 & 10.4 \\
 & weak-weak & fewest & Ministral-8B & 50.8 & Llama-3.2-3B & 39.6 & 11.2 & 7.4 \\
\midrule
Knowledge & strong-strong & most & Qwen3-32B & 69.8 & Mistral-Small-24B & 65.2 & 4.5 & 8.2 \\
 & strong-strong & fewest & Qwen2.5-72B & 72.2 & Mistral-Small-24B & 65.2 & 7.0 & 6.0 \\
 & strong-middle & most & Mistral-Small-24B & 65.2 & Qwen3-8B & 63.9 & 1.4 & 10.3 \\
 & strong-middle & fewest & Mistral-Small-24B & 65.2 & Qwen2.5-7B & 56.9 & 8.4 & 8.0 \\
 & strong-weak & most & Qwen3-32B & 69.8 & Qwen2.5-3B & 42.8 & 26.9 & 5.5 \\
 & strong-weak & fewest & Mistral-Small-24B & 65.2 & Llama-3.2-3B & 38.4 & 26.8 & 5.4 \\
 & middle-middle & most & Qwen3-4B & 59.1 & Qwen2.5-7B & 56.9 & 2.2 & 11.3 \\
 & middle-middle & fewest & Qwen3-8B & 63.9 & Qwen2.5-7B & 56.9 & 7.0 & 9.0 \\
 & middle-weak & most & Qwen2.5-7B & 56.9 & Qwen2.5-3B & 42.8 & 14.0 & 7.8 \\
 & middle-weak & fewest & Qwen3-8B & 63.9 & Llama-3.2-3B & 38.4 & 25.4 & 5.9 \\
 & weak-weak & most & Qwen2.5-3B & 42.8 & Ministral-8B & 40.5 & 2.4 & 13.6 \\
 & weak-weak & fewest & Qwen2.5-3B & 42.8 & Llama-3.2-3B & 38.4 & 4.4 & 12.9 \\
\midrule
Code & strong-strong & most & Qwen3-32B & 43.8 & Mistral-Small-24B & 43.3 & 0.5 & 8.3 \\
 & strong-strong & fewest & Seed-Coder-8B & 43.9 & Mistral-Small-24B & 43.3 & 0.6 & 7.3 \\
 & strong-middle & most & Mistral-Small-24B & 43.3 & Qwen3-8B & 42.4 & 0.8 & 9.6 \\
 & strong-middle & fewest & Mistral-Small-24B & 43.3 & Ministral-8B & 37.5 & 5.8 & 6.6 \\
 & strong-weak & most & Qwen3-32B & 43.8 & Qwen2.5-7B & 33.9 & 9.9 & 6.5 \\
 & strong-weak & fewest & Seed-Coder-8B & 43.9 & Llama-3.2-3B & 24.6 & 19.2 & 3.7 \\
 & middle-middle & most & Qwen3-4B & 38.0 & Ministral-8B & 37.5 & 0.5 & 10.6 \\
 & middle-middle & fewest & Qwen2.5-Coder-7B & 40.7 & Ministral-8B & 37.5 & 3.2 & 7.2 \\
 & middle-weak & most & Qwen3-4B & 38.0 & Qwen2.5-7B & 33.9 & 4.1 & 9.5 \\
 & middle-weak & fewest & Qwen3-8B & 42.4 & Llama-3.2-3B & 24.6 & 17.8 & 3.7 \\
 & weak-weak & most & Qwen2.5-7B & 33.9 & Qwen2.5-3B & 31.1 & 2.8 & 10.7 \\
 & weak-weak & fewest & Qwen2.5-7B & 33.9 & Llama-3.2-3B & 24.6 & 9.3 & 7.9 \\
\midrule
Reasoning & strong-strong & most & Qwen3-8B & 73.8 & Mistral-Small-24B & 71.6 & 2.2 & 9.2 \\
 & strong-strong & fewest & Qwen3-32B & 77.8 & Qwen3-8B & 73.8 & 4.0 & 5.5 \\
 & strong-middle & most & Mistral-Small-24B & 71.6 & Qwen3-4B & 67.6 & 4.0 & 10.7 \\
 & strong-middle & fewest & Qwen3-8B & 73.8 & Qwen3-4B & 67.6 & 6.2 & 5.7 \\
 & strong-middle & fewest & Qwen3-32B & 77.8 & Qwen3-4B & 67.6 & 10.2 & 5.7 \\
 & strong-weak & most & Mistral-Small-24B & 71.6 & Qwen2.5-3B & 55.2 & 16.4 & 9.2 \\
 & strong-weak & fewest & Qwen3-32B & 77.8 & Llama-3.2-3B & 43.7 & 34.1 & 4.9 \\
 & middle-middle & only & Qwen3-4B & 67.6 & Qwen2.5-7B & 61.7 & 5.9 & 9.1 \\
 & middle-weak & most & Qwen2.5-7B & 61.7 & Qwen2.5-3B & 55.2 & 6.5 & 11.2 \\
 & middle-weak & fewest & Qwen3-4B & 67.6 & Llama-3.2-3B & 43.7 & 23.9 & 6.9 \\
 & weak-weak & most & Qwen2.5-3B & 55.2 & Ministral-8B & 51.8 & 3.3 & 12.7 \\
 & weak-weak & fewest & Qwen2.5-3B & 55.2 & Llama-3.2-3B & 43.7 & 11.4 & 10.8 \\
    \bottomrule
  \end{tabular}
  \caption{The 48 model pairs. Tiers and the headroom end (most or fewest questions that only the weaker model answers
  correctly) come from the development questions; accuracy, gap and headroom $h$ (\%) are over all questions used in
  the analyses (2,500 in math and knowledge, 832 in code, 901 in reasoning). In math, the Qwen3-4B and Qwen3-8B pair
  uses their answers from an earlier run on the test questions, so Qwen3-8B's accuracy differs by 0.2 points from
  its other pairs.}
  \label{tab:pairs}
\end{table*}

\section{Prompts}
\label{app:prompts}
Every call sends a system message and a user message. Each domain has one solve prompt and one review prompt, and
reasoning uses the knowledge prompts unchanged. Both models of a pair answer alone with the same solve prompt. In the
templates below, \texttt{\{\{problem\}\}}, \texttt{\{\{question\}\}} and \texttt{\{\{options\}\}} are replaced by the
question, and \texttt{\{\{y0\}\}} by G's complete response as plain text. Options are listed one per line as
``A.~option''. A math problem is the MATH problem statement and a code task is the BigCodeBench-Instruct prompt. A
ReClor question is the passage, a blank line and the question. A FOLIO question gives the premises and the conclusion
and asks ``Based only on the premises, is the conclusion true, false, or uncertain?'', with the options True, False
and Uncertain. GPQA lists the correct answer first, so each question's options are shuffled with a fixed permutation.

\begin{lstlisting}[title={Math, solve}]
System:
Solve the mathematical problem provided by the user. Use mathematically sound reasoning and calculations to produce a complete solution.
End with exactly one final answer in LaTeX boxed form: \boxed{...}.

User template:
Problem:
{{problem}}
\end{lstlisting}

\begin{lstlisting}[title={Math, review}]
System:
Review the proposed mathematical solution for correctness.
The proposed solution may be correct or incorrect. Carefully check the reasoning, calculations, assumptions, and final answer.
If the proposed solution is correct, retain it. If it contains any mathematical error, correct the error and produce a correct solution.
Return a complete final solution to the original problem. End with exactly one final answer in LaTeX boxed form: \boxed{...}.

User template:
Problem:
{{problem}}

Proposed solution:
{{y0}}
\end{lstlisting}

\begin{lstlisting}[title={Knowledge and reasoning, solve}]
System:
Answer the multiple-choice question provided by the user. Think step by step before answering.
The last line of your response must be of the form "Answer: X", where X is the letter of exactly one option.

User template:
Question:
{{question}}

Options:
{{options}}
\end{lstlisting}

\begin{lstlisting}[title={Knowledge and reasoning, review}]
System:
Review the proposed solution to the multiple-choice question for correctness.
The proposed solution may be correct or incorrect. Carefully check its facts, its reasoning, and its final choice.
If the proposed solution is correct, retain its answer. If it contains any error, correct the error and produce a correct solution.
Return a complete final solution to the original question. The last line of your response must be of the form "Answer: X", where X is the letter of exactly one option.

User template:
Question:
{{question}}

Options:
{{options}}

Proposed solution:
{{y0}}
\end{lstlisting}

\begin{lstlisting}[title={Code, solve}]
System:
Complete the Python programming task provided by the user. Use sound reasoning to produce a correct and complete solution.
End with exactly one complete, self-contained Python code block (```python ... ```) that includes all required imports and the requested function definition.

User template:
Task:
{{problem}}
\end{lstlisting}

\begin{lstlisting}[title={Code, review}]
System:
Review the proposed solution to the Python programming task for correctness.
The proposed solution may be correct or incorrect. Carefully check its logic, edge cases, use of library APIs, and whether it meets every requirement of the task.
If the proposed solution is correct, retain it. If it contains any error, correct the error and produce a correct solution.
Return a complete final solution to the original task. End with exactly one complete, self-contained Python code block (```python ... ```) that includes all required imports and the requested function definition.

User template:
Task:
{{problem}}

Proposed solution:
{{y0}}
\end{lstlisting}

\begin{lstlisting}[title={Math, commit-first review (Appendix~\ref{app:checks} only)}]
System:
Solve the mathematical problem provided by the user. A proposed solution is also given, but do not use it until you have committed to your own answer.
First, solve the problem independently and state your answer on its own line in the form "My answer: ..." (without \boxed). Only then read the proposed solution and compare it with your own. Either one may be correct or incorrect: carefully check the reasoning, calculations, assumptions, and final answer of both, and decide which final answer is correct.
Return a complete final solution to the original problem. Do not use \boxed before the end. End with exactly one final answer in LaTeX boxed form: \boxed{...}.

User template: as for math review.
\end{lstlisting}

\begin{lstlisting}[title={Knowledge, commit-first review (Appendix~\ref{app:checks} only)}]
System:
Answer the multiple-choice question provided by the user. A proposed solution is also given, but do not use it until you have committed to your own answer.
First, think step by step and answer the question independently, and state your answer on its own line in the form "My answer: X", where X is the letter of one option. Only then read the proposed solution and compare it with your own. Either one may be correct or incorrect: carefully check the facts, the reasoning, and the final choice of both, and decide which option is correct.
Return a complete final solution to the original question. The last line of your response must be of the form "Answer: X", where X is the letter of exactly one option.

User template: as for knowledge review.
\end{lstlisting}

\begin{lstlisting}[title={Math, both answers shown (Appendix~\ref{app:checks} only)}]
System:
Solve the mathematical problem provided by the user. Two responses to the problem are also given, labelled Response A and Response B. Which response gives the correct final answer to the mathematical problem? It is possible that neither does. Carefully check the reasoning, calculations, assumptions, and final answer of both.
Return a complete final solution to the original problem. Do not use \boxed before the end. End with exactly one final answer in LaTeX boxed form: \boxed{...}.

User template:
Problem:
{{problem}}

Response A:
{{response_a}}

Response B:
{{response_b}}
\end{lstlisting}

\begin{lstlisting}[title={Knowledge, both answers shown (Appendix~\ref{app:checks} only)}]
System:
Answer the multiple-choice question provided by the user. Two responses to the question are also given, labelled Response A and Response B. Which response gives the correct final answer to the multiple-choice question? It is possible that neither does. Carefully check the facts, the reasoning, and the final choice of both.
Return a complete final solution to the original question. The last line of your response must be of the form "Answer: X", where X is the letter of exactly one option.

User template:
Question:
{{question}}

Options:
{{options}}

Response A:
{{response_a}}

Response B:
{{response_b}}
\end{lstlisting}

\section{Selector and JEV}
\label{app:selector}
The selector sees, for each output, token log-probabilities and entropy, length, and whether an answer can be parsed;
for the pair, whether the two answers agree; and for the question, its length and, where the dataset provides one,
its subject. Its probability estimate is the average of an L2 logistic regression and a small gradient-boosted tree.
Math and knowledge use their 1,500 held-out test questions; code and reasoning have fewer questions and use five-fold
cross-fitting. All numbers in Section~\ref{sec:cost} use these questions, so for math and knowledge they differ
slightly from Section~\ref{sec:results}, which uses all 2,500.

JEV reads the question and both full answers and returns the probability that each answer is right. A judge can
favour an answer for its position \citep{wang2024fair}, so we call it twice, once with each answer first, and average
the two. Where the two answers agree there is nothing to pick and no call is made. If the call fails or the two
averaged probabilities are equal, we return the stronger model's answer, as the selector does. Every response
reports the model version jev-1.13.0.

\paragraph{A third call.}
A reviewer could also answer the question alone first and then, in a further call, read its own answer and G's and
choose between them. This design picks between two independent answers, as our selector does, but adds a third LLM
call, while we compare ways of using a second model at the same number of calls, which \citet{huang2024cannot} argue is
crucial for methods that use several model responses. The last call can also be anchored to either answer:
committing to an answer first does not remove the anchor but moves it from the candidate to the judge's own answer
\citep{gozel2026commitfirst}, and a model that has answered first is still misled by its peers \citep{qu2026mislead}.
We ran this design on the held-out math and knowledge questions (Appendix~\ref{app:checks}, ``Shown both answers''):
with three calls it stays below the selector with two. Writing its own answer before reading G's
within one call \citep{zhou2026convincing} keeps the number of calls at two, so we ran this variant on four
directions: the reviewer still kept only 55\% of its right answers on questions only it got right alone, against
61\% with our prompt (Appendix~\ref{app:checks}). JEV is a third call of a different kind: it writes no
text, only label probabilities, at a small fraction of a reasoning judge's fee and latency \citep{li2026jev}. In our
runs it cost \$0.04--\$0.21 per 1,000 questions, and a sample of 196 calls took a median of 0.28 seconds. We report
it only as a reference, not as a way of using the second model.

\paragraph{Selector details.}
The selector estimates, for each of the two answers, the probability that it is correct and returns the answer with
the higher estimate; a tie goes to the stronger model. Each estimate averages two models over five random seeds: an
L2 logistic regression on that answer's own features and the question features, and a gradient-boosted tree on all
features of Table~\ref{tab:features}, including the other answer and the pair. Hyperparameters are chosen by
five-fold cross-validation on the training questions with log loss (logistic regression: $C \in \{0.01, 0.1, 1,
10\}$; tree: depth 2 or 3, learning rate 0.05 or 0.1, 100 or 300 iterations). Missing values are replaced by the
median plus an indicator for the logistic regression and left to the tree. Each pair has its own selector. In math
and knowledge it is trained on that pair's 1,000 development questions; in code and reasoning, cross-fitting nests this search inside each of the
five outer folds. Every feature is computed from the question and the two outputs with their token probabilities; the
reference answer, the grades and the question identifiers never enter. Code has the most features because a program's
structure can be read without running it. Reasoning reuses the knowledge selector unchanged, so its 18 subject and
source indicators, defined for MMLU-Pro, are always zero there.

\begin{table*}[t]
  \centering
  \small
  \begin{tabularx}{\textwidth}{@{}l>{\raggedright\arraybackslash}X>{\raggedright\arraybackslash}X>{\raggedright\arraybackslash}X@{}}
    \toprule
    & Math & Knowledge and reasoning & Code \\
    \midrule
    \multicolumn{4}{@{}l}{\emph{Each answer}} \\
    Basic & 7: mean log-probability and largest logit over the output and over math expressions; answer parseable;
      question and output length in characters
      & 22: answer parseable; number of options; question length; prompt and output tokens; truncated; 12 token
      statistics over the output; log-probability, entropy, margin and largest logit at the answer letter
      & 17: question length; prompt and output tokens; truncated; number of code blocks; 12 token statistics over
      the output \\
    Token spans & 33: 11 token statistics each over the output, the math expressions and the boxed answer
      & -- & 22: 11 token statistics each over the code and the prose \\
    Content & 4: output tokens; truncated; number of boxed answers; self-correction words
      & 16: spread of the answer-letter probability over the options (7); letters mentioned, answer lines,
      whether the first letter is the last, hedging, elimination and self-correction words (6); digits, numbers and
      equals signs (3)
      & 26: program structure from the syntax tree (16); use of the required libraries and of the exceptions the
      task asks for (3); undefined names, unseeded randomness, prints, main guard, file access (5); output
      characters and self-correction words (2) \\
    Total & 44 & 38 & 65 \\
    \midrule
    Question & 8: difficulty level; 7 subject indicators
      & 18: 14 subject and 4 source indicators (zero in reasoning)
      & 3: required libraries; whether and how many exceptions the task asks for \\
    Pair & 19: answers equivalent; overlap of math expressions; same answer type; parseable and truncated in one
      answer only (2); differences of 14 answer features
      & 21: same letter; both parseable; parseable and truncated in one answer only (2); divergence of the two
      option distributions and how each rates the other's letter (5); differences of 12 answer features
      & 25: identical program and syntax tree; program similarity; overlap of imports, calls and raised exceptions
      (6); four properties in one program only; differences of 15 answer features \\
    \midrule
    Logistic regression & 52 & 56 & 68 \\
    Gradient-boosted tree & 115 & 115 & 158 \\
    \bottomrule
  \end{tabularx}
  \caption{Selector features per domain, with counts. The logistic regression sees one answer's features and the
  question features; the tree sees both answers, the question and the pair. Token statistics are the mean, minimum,
  10th percentile and standard deviation of the log-probability of each output token, the mean, 90th percentile and
  maximum entropy, the mean and 10th percentile of the margin between the two most likely tokens, and the share and
  longest run of tokens with probability below 0.5; the 12 statistics add the mean largest logit.}
  \label{tab:features}
\end{table*}

\section{Cost with output tokens weighted four times}
\label{app:cost}
Section~\ref{sec:cost} counts input and output tokens alike. Generating a token usually costs more than reading one,
so Table~\ref{tab:cost4} repeats the cost ratios with $c' = N\,(t_{\text{in}} + 4\,t_{\text{out}})$. Accuracy does
not depend on the cost measure and is unchanged. Review becomes cheaper relative to the stronger model alone, 2.2 and
2.0 times instead of 2.8 and 2.6, since much of a reviewer's cost is reading G's whole answer, while the selector's
cost hardly moves (1.9 times). Against review, the selector then costs 0.89 times as much as the stronger model
reviewing and 0.98 times as much as the weaker one reviewing, and it is cheaper in 34 and 28 of the 48 pairs instead
of all 48; against the weaker model reviewing, it costs slightly more in math, knowledge and reasoning. With output
weighted this heavily, choosing between two answers keeps its accuracy advantage over review and still costs less on
average.

\begin{table*}[t]
  \centering
  \small
  \begin{tabular}{@{}lcccccccc@{}}
    \toprule
    & Main & \multicolumn{5}{c}{Output $\times 4$} & \multicolumn{2}{c}{Pairs cheaper} \\
    \cmidrule(lr){2-2} \cmidrule(lr){3-7} \cmidrule(l){8-9}
    & All & All & Math & Knowl. & Code & Reas. & Main & $\times 4$ \\
    \midrule
    \multicolumn{9}{@{}l}{\emph{Relative to the stronger model alone}} \\
    Stronger reviews & 2.76 & 2.24 & 2.17 & 1.92 & 3.05 & 1.83 & & \\
    Weaker reviews & 2.56 & 2.05 & 2.03 & 1.75 & 2.66 & 1.75 & & \\
    Selector & 1.93 & 1.94 & 2.12 & 1.77 & 2.11 & 1.77 & & \\
    \midrule
    \multicolumn{9}{@{}l}{\emph{Selector relative to review}} \\
    vs.\ stronger reviews & 0.69 & 0.89 & 0.98 & 0.92 & 0.69 & 0.96 & 48 & 34 \\
    vs.\ weaker reviews & 0.82 & 0.98 & 1.05 & 1.02 & 0.83 & 1.02 & 48 & 28 \\
    \bottomrule
  \end{tabular}
  \caption{Cost ratios under the main cost $c = N\,(t_{\text{in}} + t_{\text{out}})$ and with output tokens weighted
  four times, $c' = N\,(t_{\text{in}} + 4\,t_{\text{out}})$: the ratio is taken per pair, then averaged with domains
  weighted equally. Pairs cheaper: of 48, the pairs where the selector costs less than review.}
  \label{tab:cost4}
\end{table*}

\section{Per-domain results and robustness}
\label{app:robustness}
\paragraph{Bootstrap.}
With the models and pairs fixed by design, the main intervals measure the uncertainty from which questions were
sampled. For one direction, we resample questions with replacement, keeping each question's paired outcomes together
(2,000 replicates, percentile 95\% intervals), within the development and test parts and, in reasoning, within each
source dataset. In each replicate, all directions of a domain share one resample of questions. The regressions on the
gap are fitted separately for the two kinds of direction, and since they compare pairs, their bootstrap also resamples
pairs within each domain.

\paragraph{Per domain.}
Table~\ref{tab:domain-rq} splits the averages and outcome levels of Section~\ref{sec:results} by domain. Review falls
below the stronger model on average in every domain and in both kinds of direction, with one exception: a weaker model
reviewing a stronger one in knowledge is level with it ($+0.1$ [$-0.3$, $+0.5$]), and five of the eight weaker-reviewer
directions beyond the stronger model are there. Code is where review does worst: a stronger reviewer lifts G by only 2.7 points,
and no weaker reviewer improves on G. Table~\ref{tab:domain-cost} gives the comparisons of Section~\ref{sec:cost} by
domain: the selector is above the stronger model in every domain except code, and above both review directions in
every domain.

\begin{table*}[t]
  \centering
  \small
  \setlength{\tabcolsep}{4pt}
  \begin{adjustbox}{max width=\textwidth}
  \begin{tabular}{@{}lllcccccllccc@{}}
    \toprule
    & \multicolumn{7}{c}{Stronger model reviews} & \multicolumn{5}{c}{Weaker model reviews} \\
    \cmidrule(lr){2-8} \cmidrule(l){9-13}
    & $\Delta_G$ & $\Delta_S$ & W & N & A & L & B & \multicolumn{2}{l}{$\Delta_S = \Delta_G$} & W & N & B \\
    \midrule
    Math & $+10.3$ {\scriptsize[$+9.9$, $+10.9$]} & $-4.5$ {\scriptsize[$-5.1$, $-3.9$]} & 0 & 1 & 9 & 2 & 0
      & \multicolumn{2}{l}{$-1.3$ {\scriptsize[$-1.5$, $-1.0$]}} & 6 & 3 & 3 \\
    Knowledge & $+9.2$ {\scriptsize[$+8.5$, $+9.8$]} & $-1.7$ {\scriptsize[$-2.3$, $-1.1$]} & 0 & 0 & 4 & 7 & 1
      & \multicolumn{2}{l}{$+0.1$ {\scriptsize[$-0.3$, $+0.5$]}} & 4 & 3 & 5 \\
    Code & $+2.7$ {\scriptsize[$+1.9$, $+3.6$]} & $-3.5$ {\scriptsize[$-4.6$, $-2.3$]} & 1 & 3 & 3 & 5 & 0
      & \multicolumn{2}{l}{$-4.0$ {\scriptsize[$-4.8$, $-3.2$]}} & 8 & 4 & 0 \\
    Reasoning & $+8.0$ {\scriptsize[$+7.1$, $+9.0$]} & $-2.7$ {\scriptsize[$-3.6$, $-1.7$]} & 0 & 1 & 6 & 5 & 0
      & \multicolumn{2}{l}{$-1.1$ {\scriptsize[$-1.7$, $-0.4$]}} & 4 & 8 & 0 \\
    \midrule
    All & $+7.6$ {\scriptsize[$+7.2$, $+7.9$]} & $-3.1$ {\scriptsize[$-3.5$, $-2.7$]} & 1 & 5 & 22 & 19 & 1
      & \multicolumn{2}{l}{$-1.6$ {\scriptsize[$-1.8$, $-1.3$]}} & 22 & 18 & 8 \\
    \bottomrule
  \end{tabular}
  \end{adjustbox}
  \caption{Section~\ref{sec:results} by domain: mean $\Delta_G$ and $\Delta_S$ (points) over the 12 directions of each
  kind, with 95\% intervals that resample questions, and the number of directions at each outcome level: worse than
  G (W), no different from G (N), above G but below the stronger model (A), level with the stronger model (L), beyond
  it (B). When the weaker model reviews, G is the stronger model, so $\Delta_S = \Delta_G$ and only W, N and B occur.
  All: domains weighted equally.}
  \label{tab:domain-rq}
\end{table*}

\begin{table*}[t]
  \centering
  \small
  \setlength{\tabcolsep}{4pt}
  \begin{adjustbox}{max width=\textwidth}
  \begin{tabular}{@{}lcccccc@{}}
    \toprule
    & \multicolumn{4}{c}{Relative to the stronger model alone} & \multicolumn{2}{c}{Selector relative to review} \\
    \cmidrule(lr){2-5} \cmidrule(l){6-7}
    & Stronger reviews & Weaker reviews & Selector & JEV & vs.\ stronger reviews & vs.\ weaker reviews \\
    \midrule
    Math & $-4.7$ {\scriptsize[$-5.4$, $-4.0$]} & $-1.4$ {\scriptsize[$-1.8$, $-1.1$]} & $+1.2$ {\scriptsize[$+0.9$, $+1.5$]}
      & $+3.7$ {\scriptsize[$+3.3$, $+4.1$]} & $+5.9$ {\scriptsize[$+5.2$, $+6.6$]} & $+2.6$ {\scriptsize[$+2.3$, $+3.0$]} \\
    Knowledge & $-1.3$ {\scriptsize[$-2.1$, $-0.6$]} & $+0.2$ {\scriptsize[$-0.3$, $+0.6$]} & $+2.1$ {\scriptsize[$+1.7$, $+2.5$]}
      & $+5.5$ {\scriptsize[$+5.0$, $+6.0$]} & $+3.4$ {\scriptsize[$+2.7$, $+4.2$]} & $+1.9$ {\scriptsize[$+1.5$, $+2.5$]} \\
    Code & $-3.5$ {\scriptsize[$-4.7$, $-2.3$]} & $-4.0$ {\scriptsize[$-4.8$, $-3.3$]} & $+0.5$ {\scriptsize[$-0.1$, $+1.0$]}
      & $+2.0$ {\scriptsize[$+1.2$, $+2.7$]} & $+3.9$ {\scriptsize[$+2.9$, $+5.1$]} & $+4.5$ {\scriptsize[$+3.6$, $+5.4$]} \\
    Reasoning & $-2.6$ {\scriptsize[$-3.7$, $-1.7$]} & $-1.1$ {\scriptsize[$-1.7$, $-0.5$]} & $+0.9$ {\scriptsize[$+0.4$, $+1.3$]}
      & $+5.1$ {\scriptsize[$+4.4$, $+5.7$]} & $+3.5$ {\scriptsize[$+2.5$, $+4.5$]} & $+1.9$ {\scriptsize[$+1.2$, $+2.6$]} \\
    \midrule
    All & $-3.0$ {\scriptsize[$-3.5$, $-2.6$]} & $-1.6$ {\scriptsize[$-1.9$, $-1.3$]} & $+1.2$ {\scriptsize[$+0.9$, $+1.4$]}
      & $+4.1$ {\scriptsize[$+3.8$, $+4.3$]} & $+4.2$ {\scriptsize[$+3.8$, $+4.6$]} & $+2.7$ {\scriptsize[$+2.4$, $+3.1$]} \\
    \bottomrule
  \end{tabular}
  \end{adjustbox}
  \caption{Section~\ref{sec:cost} by domain: accuracy differences (points) with 95\% intervals that resample
  questions. Math and knowledge use the 1,500 held-out questions, so these numbers and intervals differ slightly from
  Table~\ref{tab:domain-rq}. All: domains weighted equally.}
  \label{tab:domain-cost}
\end{table*}

\paragraph{Leaving out one model.}
Models recur across pairs, so one model could drive an average. We repeat each average after removing every
direction that involves one of the 12 models, and Table~\ref{tab:robust} gives the range over the 12 repetitions. No
average changes sign.

\paragraph{All completed pairs.}
Besides the 48 pairs, we ran review on further pairs during the study, some in one direction only, including the
four original pairs with Qwen2.5-72B (Appendix~\ref{app:pairs}): 131 review directions in all. These never enter the
main analysis. Pooled over both kinds of direction, the averages over all of them stay close to those over the 48
pairs (Table~\ref{tab:robust}).

\begin{table*}[t]
  \centering
  \small
  \begin{tabular}{@{}lccc@{}}
    \toprule
    & 48 pairs & Leaving out one model & All completed pairs \\
    & (96 directions) & (range of the mean) & (131 directions) \\
    \midrule
    \multicolumn{4}{@{}l}{\emph{Both kinds of direction pooled}} \\
    Weaker reviews $-$ stronger reviews & $+1.5$ [$+1.1$, $+1.9$] & $+1.2$ to $+1.9$ & $+1.7$ [$+1.4$, $+2.1$] \\
    $\Delta_G$ & $+3.0$ [$+2.7$, $+3.3$] & $+2.5$ to $+3.5$ & $+3.2$ [$+3.0$, $+3.5$] \\
    $\Delta_S$ & $-2.3$ [$-2.6$, $-2.0$] & $-2.7$ to $-1.6$ & $-2.5$ [$-2.8$, $-2.2$] \\
    Net change where both agree & $+0.2$ [$+0.0$, $+0.4$] & $+0.1$ to $+0.4$ & $+0.2$ [$+0.0$, $+0.4$] \\
    Drop in disagreement with G & $+14.6$ [$+14.1$, $+15.1$] & $+13.3$ to $+16.1$ & $+14.7$ [$+14.2$, $+15.1$] \\
    Oracle over the stronger model & $+7.5$ [$+7.2$, $+7.8$] & $+7.0$ to $+8.0$ & $+7.3$ [$+7.1$, $+7.6$] \\
    \midrule
    \multicolumn{4}{@{}l}{\emph{By kind of direction}} \\
    $\Delta_G$, stronger reviews & $+7.6$ [$+7.2$, $+7.9$] & $+6.0$ to $+8.6$ & \\
    $\Delta_S$, stronger reviews & $-3.1$ [$-3.5$, $-2.7$] & $-3.6$ to $-2.2$ & \\
    $\Delta_S = \Delta_G$, weaker reviews & $-1.6$ [$-1.8$, $-1.3$] & $-1.9$ to $-1.1$ & \\
    Net change where both agree, stronger reviews & $+0.9$ [$+0.7$, $+1.2$] & $+0.7$ to $+1.3$ & \\
    Net change where both agree, weaker reviews & $-0.5$ [$-0.7$, $-0.3$] & $-0.6$ to $-0.4$ & \\
    Drop in disagreement with G, stronger reviews & $+10.7$ [$+10.2$, $+11.2$] & $+9.4$ to $+11.7$ & \\
    Drop in disagreement with G, weaker reviews & $+18.5$ [$+18.0$, $+19.0$] & $+16.6$ to $+20.5$ & \\
    \bottomrule
  \end{tabular}
  \caption{Robustness of the main averages (points, domains weighted equally, 95\% intervals that resample
  questions). Leaving out one model removes every direction involving it; the column gives the lowest and highest mean
  over the 12 models. All completed pairs adds every other pair we ran review on; it is computed with both kinds of
  direction pooled only.}
  \label{tab:robust}
\end{table*}

\paragraph{Resampling model pairs.}
The intervals in the main text keep the evaluated pairs fixed and resample questions. Table~\ref{tab:pairs-resampled}
also resamples model pairs within each domain, keeping the two directions of a pair as one unit, along with the questions. The
intervals widen, since pairs differ from one another, but every average keeps its sign. The one that then reaches
below zero is the oracle after the stronger model reviews, which Section~\ref{sec:rq2} already describes as almost
no room left. Models recur across pairs, so pairs are not fully independent and this is a robustness check rather
than an interval over a population of model pairs.

\begin{table*}[t]
  \centering
  \small
  \begin{tabular}{@{}lccc@{}}
    \toprule
    & Mean & Questions & Pairs and questions \\
    \midrule
    $\Delta_S$, stronger reviews & $-3.1$ & [$-3.5$, $-2.7$] & [$-4.1$, $-2.0$] \\
    $\Delta_S$, weaker reviews & $-1.6$ & [$-1.8$, $-1.3$] & [$-2.4$, $-0.8$] \\
    Oracle over the stronger model & $+7.5$ & [$+7.2$, $+7.8$] & [$+6.7$, $+8.3$] \\
    Oracle after review, stronger reviews & $+0.6$ & [$+0.2$, $+1.0$] & [$-0.5$, $+1.6$] \\
    Oracle after review, weaker reviews & $+2.8$ & [$+2.6$, $+2.9$] & [$+2.3$, $+3.3$] \\
    Net change where both agree, stronger reviews & $+0.9$ & [$+0.7$, $+1.2$] & [$+0.5$, $+1.4$] \\
    Net change where both agree, weaker reviews & $-0.5$ & [$-0.7$, $-0.3$] & [$-0.9$, $-0.1$] \\
    Drop in disagreement with G & $+14.6$ & [$+14.1$, $+15.1$] & [$+13.2$, $+16.0$] \\
    Selector $-$ stronger alone & $+1.2$ & [$+0.9$, $+1.4$] & [$+0.8$, $+1.6$] \\
    JEV $-$ stronger alone & $+4.1$ & [$+3.8$, $+4.3$] & [$+3.4$, $+4.8$] \\
    Selector $-$ stronger reviews & $+4.2$ & [$+3.8$, $+4.6$] & [$+3.3$, $+5.1$] \\
    Selector $-$ weaker reviews & $+2.7$ & [$+2.4$, $+3.1$] & [$+2.0$, $+3.5$] \\
    \bottomrule
  \end{tabular}
  \caption{Main averages (points, domains weighted equally) with 95\% intervals that resample questions only, as in
  the main text, or model pairs and questions.}
  \label{tab:pairs-resampled}
\end{table*}

\section{Further checks}
\label{app:checks}

\paragraph{All eight cells.}
Table~\ref{tab:eight-cells} splits the questions by whether G alone, V alone and the final answer are right. Two rows
exist only because of review: questions both models get wrong that review gets right, and questions both get right
that review gets wrong. A choice between the two independent answers, such as the selector of
Section~\ref{sec:cost}, can produce neither. When the stronger model reviews, these rows hold 2.9\% and 1.9\% of the
questions, while the reviewer's own right answers that review takes up hold 8.4\%, and those it loses 9.8\%.

\begin{table*}[t]
  \centering
  \small
  \setlength{\tabcolsep}{3pt}
  \begin{tabularx}{\textwidth}{@{}ccc >{\raggedright\arraybackslash}X rrrrr rrrrr@{}}
    \toprule
    & & & & \multicolumn{5}{c}{Stronger model reviews} & \multicolumn{5}{c}{Weaker model reviews} \\
    \cmidrule(lr){5-9} \cmidrule(l){10-14}
    G & V & Final & What happens & Math & Know. & Code & Reas. & All & Math & Know. & Code & Reas. & All \\
    \midrule
    \checkmark & \checkmark & \checkmark & Both right alone; review keeps it.
      & 49.5 & 39.9 & 24.6 & 47.5 & 40.4 & 49.4 & 40.5 & 23.8 & 48.1 & 40.4 \\
    \checkmark & \checkmark & $\times$ & Both right alone; review breaks it. Harm only review can do: a choice between
      the two answers cannot lose these.
      & 0.9 & 2.0 & 2.3 & 2.6 & 1.9 & 1.0 & 1.4 & 3.1 & 2.1 & 1.9 \\
    \checkmark & $\times$ & \checkmark & Only G right; review keeps G's answer.
      & 4.7 & 6.1 & 6.3 & 6.0 & 5.8 & 18.6 & 17.0 & 10.4 & 16.8 & 15.7 \\
    \checkmark & $\times$ & $\times$ & Only G right; review loses it (the G-only part of $\Delta_G$).
      & 0.7 & 2.6 & 1.4 & 2.5 & 1.8 & 1.6 & 2.5 & 3.4 & 2.4 & 2.5 \\
    $\times$ & \checkmark & \checkmark & Only V right; review takes up V's answer (the V-only part of $\Delta_G$).
      Capability transfer: V alone answers these right too.
      & 9.5 & 10.2 & 4.0 & 9.9 & 8.4 & 0.5 & 2.3 & 1.1 & 1.7 & 1.4 \\
    $\times$ & \checkmark & $\times$ & Only V right; review loses V's right answer, most often to G's wrong one.
      A choice between the two answers can keep these.
      & 10.7 & 9.3 & 9.9 & 9.2 & 9.8 & 4.8 & 6.4 & 6.5 & 6.8 & 6.1 \\
    $\times$ & $\times$ & \checkmark & Both wrong alone; review gets it right. Value only review can create: neither
      model alone nor a choice between the two answers gets these.
      & 2.4 & 3.5 & 2.4 & 3.2 & 2.9 & 0.8 & 1.7 & 1.4 & 1.7 & 1.4 \\
    $\times$ & $\times$ & $\times$ & Both wrong alone and after review.
      & 21.7 & 26.4 & 49.2 & 19.0 & 29.1 & 23.2 & 28.2 & 50.2 & 20.5 & 30.6 \\
    \bottomrule
  \end{tabularx}
  \caption{All eight cells: the share of questions (\%) by whether G alone, V alone and the final answer are right,
  averaged over the 12 directions of each kind in a domain; All weights the domains equally. The three parts of
  $\Delta_G$ in Section~\ref{sec:review} are the V-only row, minus the G-only row, plus the both-wrong-then-right row
  minus the both-right-then-wrong row. Columns may not sum to 100 because of rounding.}
  \label{tab:eight-cells}
\end{table*}

\paragraph{Does the reviewer follow G?}
When the stronger model reviews, its final answer is the same as G's on 57.7\% [57.2, 58.2] of the questions of
Section~\ref{sec:results}, against 41.8\% [41.2, 42.3] for its own solo answer; when the weaker model reviews, the
final answer is G's on 69.1\% [68.7, 69.4]. Answers count as the same when the extracted answers are identical (the
option letter in knowledge and reasoning, the whole final program in code). Of the right answers a reviewer loses on
questions only it got right alone, most end on G's wrong answer: 53.3\% [52.2, 54.5] when the stronger model reviews
and 65.0\% [63.7, 66.3] when the weaker one does. In code, the reviewer's own program never equals G's, yet its final
program does on 21\% of the questions.

\paragraph{When does the reviewer keep its own answer?}
On the 21,247 questions where G was wrong and V right alone, we relate whether the final answer stays right to the
reviewer's strength relative to G and to features of the two solo answers, compared within each direction (logistic
regression with direction fixed effects; Table~\ref{tab:keep}). Across directions, the further a stronger reviewer is
above G, the more of these answers it keeps ($+13.8$ points per 10 points of gap, $r=0.74$), and the further a weaker
reviewer is below G, the fewer ($-4.6$ points, $r=-0.41$). Within a direction, the reviewer keeps fewer of them the
longer its own solution and more the higher its mean maximum logit. G's answer shows no association: the intervals
for its length, maximum logit and log-probability all contain 1, except that a weaker reviewer keeps more when G's
output was cut off at the token limit. This matches the thinking check below, where the reviewer stops following a
weaker G but still follows a stronger one. These are correlations, not causal effects: we vary none of these
features, and the questions are selected because the reviewer was right alone.

\begin{table*}[t]
  \centering
  \small
  \setlength{\tabcolsep}{4pt}
  \begin{tabular}{@{}lllll@{}}
    \toprule
    & \multicolumn{2}{c}{Stronger model reviews (15,364)} & \multicolumn{2}{c}{Weaker model reviews (5,883)} \\
    \cmidrule(lr){2-3} \cmidrule(l){4-5}
    Feature & Odds ratio & Kept (\%), Q1 to Q4 & Odds ratio & Kept (\%), Q1 to Q4 \\
    \midrule
    V's solo output length & \textbf{0.75 {\scriptsize[0.71, 0.79]}} & 54 \,/\, 51 \,/\, 47 \,/\, 39 & \textbf{0.76 {\scriptsize[0.69, 0.83]}} & 23 \,/\, 21 \,/\, 19 \,/\, 15 \\
    V's mean maximum logit & \textbf{1.16 {\scriptsize[1.10, 1.22]}} & 43 \,/\, 46 \,/\, 51 \,/\, 52 & \textbf{1.24 {\scriptsize[1.11, 1.38]}} & 16 \,/\, 18 \,/\, 21 \,/\, 23 \\
    V's mean log-probability & 1.06 {\scriptsize[1.00, 1.13]} & 47 \,/\, 48 \,/\, 48 \,/\, 48 & 1.04 {\scriptsize[0.93, 1.18]} & 18 \,/\, 17 \,/\, 20 \,/\, 21 \\
    G's output length & 1.00 {\scriptsize[0.93, 1.07]} & 52 \,/\, 49 \,/\, 46 \,/\, 45 & 0.94 {\scriptsize[0.84, 1.06]} & 21 \,/\, 20 \,/\, 19 \,/\, 18 \\
    G's mean maximum logit & 0.96 {\scriptsize[0.91, 1.01]} & 48 \,/\, 47 \,/\, 49 \,/\, 48 & 0.96 {\scriptsize[0.87, 1.06]} & 20 \,/\, 17 \,/\, 19 \,/\, 21 \\
    G's mean log-probability & 0.96 {\scriptsize[0.90, 1.02]} & 49 \,/\, 49 \,/\, 47 \,/\, 47 & 1.02 {\scriptsize[0.91, 1.14]} & 19 \,/\, 18 \,/\, 19 \,/\, 21 \\
    G's output cut off at the limit & 1.06 {\scriptsize[0.99, 1.13]} & no 47 \,/\, yes 55 & \textbf{1.17 {\scriptsize[1.06, 1.30]}} & no 19 \,/\, yes 28 \\
    \bottomrule
  \end{tabular}
  \caption{When the reviewer keeps its own right answer, on questions where G was wrong and V right alone (number of
  questions in parentheses). Odds ratio: change in the odds that the final answer stays right per standard deviation
  of the feature, from one logistic regression per kind of direction with all seven features and direction fixed
  effects; features are standardised within each direction, since raw logits differ in scale across models; 95\%
  intervals clustered by question; bold intervals exclude 1. Kept: share of these questions whose final answer stays
  right, by quartile of the feature within each direction (Q1 lowest). Correlations, not causal effects.}
  \label{tab:keep}
\end{table*}

\paragraph{Self-review as a control.}
Without external feedback, models do not improve their own reasoning and can even lose accuracy
\citep{huang2024cannot}. This makes self-review a control for following G: the same reviewer, prompt and questions,
with the reviewer's own solo answer shown in place of G's. If the review step itself made reviewers drop right
answers, they would drop them under self-review too. On 1,000 MATH questions used during development, with the same
prompt, Qwen3-8B reviewing its own answers kept 253 of its 254 right answers that Ministral-8B had missed, but kept 157
when it reviewed Ministral-8B; it got 783 questions right alone, 788 under self-review and 743 after reviewing
Ministral-8B. Ministral-8B kept all 22 of its right answers that Qwen3-8B had missed under self-review and none after
reviewing Qwen3-8B, and 20 of the 22 it lost became Qwen3-8B's answer.
On the first 500 held-out questions of math
and knowledge, we ran self-review for the four models that review in
the design directions (11 directions; two more take the reviewer's solo answers from a different run and are left
out). Summed over the directions, of the right answers only the reviewer had, self-review kept 467 of 468 in math
and 449 of 493 in knowledge, while reviewing G kept 306 and 218. Each model's accuracy under self-review stayed within
five questions of its accuracy alone (414 and 416, 407 and 405 in math; 318 and 313, 281 and 283 in knowledge).
Self-review changes the prompt and the context just as reviewing G does, and its questions are selected the same way,
yet it keeps 99.8\% of these right answers in math and 91\% in knowledge, against 65\% and 44\% after reviewing G. So
the new prompt and context, and regression to the mean on questions chosen because the reviewer got them right,
explain at most a small part of the losses after reviewing G. The rest comes from anchoring on G's answer, and most of
the lost answers become G's answer (see above). Self-review may also keep answers because models prefer their own
outputs \citep{panickssery2024recognize}; either way, the reviewer keeps the answer it is shown, and the losses come
from being shown G's.

\paragraph{A prompt without the instruction to retain.}
Our review prompt tells the reviewer to retain a proposed solution that is correct, which makes G's answer the
default, so the reviewer might follow G because the prompt tells it to. We reran four directions on the same 500
held-out questions as the self-review check with only the review prompt changed: the reviewer first solves the
question and states its own answer on a line of its own, then reads G's solution, told that either may be right or
wrong, and returns a final solution (Appendix~\ref{app:prompts}). The instruction to retain is gone. This is the
one-call commit-first protocol of \citet{zhou2026convincing}. The directions are Qwen3-8B reviewing Llama-3.1-8B and
Qwen3-32B in math, and Qwen3-8B reviewing Llama-3.2-3B and Qwen2.5-7B reviewing Qwen3-4B in knowledge, two where the
stronger model reviews and two where the weaker one does. Of the right answers only the reviewer had, the new prompt
kept 231 of 421 (55\%), against 256 (61\%) with our prompt and 404 (96\%) under self-review; per direction, 143, 4,
69 and 15, against 157, 3, 79 and 17. The change over the stronger model alone moved from $-6.8$, $+0.6$, $-7.8$ and
$+2.0$ points to $-9.2$, $+1.0$, $-5.4$ and $+2.2$. Whenever the committed answer was right and G's was wrong, the
final answer stayed right (143 of 144 questions). The answers are lost earlier: with G's solution already in its
input, the reviewer often commits to an answer that is not its own solo answer, which in knowledge it gives on only
64\% and 66\% of the questions with a committed answer (in math the reviewer wrote a committed answer on only 211 and
310 of the 500, so there the variant was often not carried out as written). Of the committed answers that differ from
the reviewer's solo answer, 270 of 479 are G's answer. Following G is not an effect of the instruction to retain.

\paragraph{Shown both answers.}
A reviewer that sees its own solo answer next to G's could keep its right answers without being misled by G. On the
1,500 held-out questions of math and knowledge, in the 48 directions of the 24 pairs there, we called each reviewer
once more with its own solo response and G's, unlabelled as Response A and Response B, asked which gives the correct
answer, possibly neither, and asked for a complete final solution (prompt in Appendix~\ref{app:prompts}; decoding as
in review). We ran both orders on every question where exactly one of the two models is right alone. Where only the
reviewer was right, it now ends right on 77\% and 79\% of the questions (its own response first, G's first), against
42\% after review. Where only G was right, it ends right on 60\% and 53\%, against 86\% after review: shown its own
wrong answer, it often returns to it. How this nets out depends on which model is stronger (Table~\ref{tab:both}).
When the stronger model reviews, it ends 2.0 [1.0, 2.9] and 1.1 [0.1, 2.0] points above the stronger model alone on
these questions, but 4.2 [3.3, 5.2] and 5.1 [4.2, 6.1] points below the selector, which uses one call fewer. When the
weaker model reviews, it ends below review itself, and 17 to 22 points below the selector. The order of the two
responses also matters: of the right answers only the reviewer had and lost in review, it recovers 54\% in both
orders. Showing the reviewer its own answer moves the anchor from G's answer to its own, as
\citet{gozel2026commitfirst} argue for commit-first judging, and helps only when the reviewer is the stronger model.

\begin{table}[t]
  \centering
  \small
  \begin{tabular}{@{}lrr@{}}
    \toprule
    & Stronger reviews & Weaker reviews \\
    \midrule
    Stronger model alone & 73.2 & 73.2 \\
    Selector & 79.4 & 79.4 \\
    Review & 56.4 & 70.8 \\
    Both shown, V's first & 75.2 & 62.0 \\
    Both shown, G's first & 74.2 & 57.7 \\
    \bottomrule
  \end{tabular}
  \caption{Accuracy (\%) on the questions where exactly one of the two models is right alone (held-out math and
  knowledge questions, 24 pairs, 9,653 per kind of direction). The two directions of a pair share these questions,
  so the stronger model alone and the selector are the same in both columns. On the other questions the stronger model
  alone and the selector are fixed, but the reviewer shown both answers was not run there. A response without an
  answer counts as wrong.}
  \label{tab:both}
\end{table}

\paragraph{Thinking on.}
All our runs keep Qwen3's thinking off. A reviewer that reasons at length before answering might check G's solution
instead of following it. We reran Qwen3-8B with thinking on, alone and reviewing Llama-3.1-8B and Qwen3-32B, on the
same 500 held-out math questions as the prompt check, with G's answers unchanged. Qwen advises against greedy decoding
with thinking on, so we sample with its recommended settings (temperature 0.6, top-p 0.95, top-k 20) and allow up to
32,768 output tokens. Thinking raises Qwen3-8B alone from 414 to 482 right answers, and review still falls below it:
by 2.8 points [1.0, 4.6] reviewing Llama-3.1-8B and 3.0 [0.8, 5.2] reviewing Qwen3-32B, mostly on questions only the
reviewer got right alone. How it loses them depends on G. Reviewing the weaker Llama-3.1-8B, it keeps 252 of 270 of
these answers (93\%, against 77\% with thinking off), and only 3 of the 18 it loses become Llama-3.1-8B's answer.
Reviewing the stronger Qwen3-32B, it now answers many more of these questions alone (64 against 23), but still loses
about as many (21 against 20), and 14 of them become Qwen3-32B's answer (17 with thinking off). Because we sample
with thinking on, a question answered right once alone is not always answered right again; this can explain losses
to other answers, but not losses to G's. So with thinking on, the reviewer appears to stop following a weaker G but
still follows a stronger one; why, and whether this holds for other models and domains, is left for future work.
Each review output grows from about 800 tokens to 2,900 and 4,000.
We also turned thinking on for both models, with Qwen3-4B, Qwen3-8B and Qwen3-32B reviewing each other on the same
500 questions. Alone, all three now get 96--98\% of them right, so these questions leave too little room to measure
what review adds: in all six directions it ends within one point of the stronger model alone. The reviewer still
keeps only 24 of the 48 right answers that only it had, against 47 of the 48 that only G had.

\paragraph{Output length.}
Few outputs stop at the 4,096-token limit: 1.7\% of review outputs when the stronger model reviews and 1.8\% when the
weaker one does, against 1.8\% for the stronger model alone and 3.3\% for the weaker one. Math has the most (4.4--4.5\%
of review outputs, 5.0--7.4\% of solo answers). Review outputs are cut off no more often than the stronger model's own
answers, so the limit is unlikely to explain why review loses right answers.

\paragraph{A selector without labels.}
Our selector is trained on development questions with known answers. We also tried a rule that needs no answers:
return the answer whose output has the higher mean token log-probability, and the stronger model's answer on ties.
It is 3.7 points [3.4, 4.1] below the stronger model alone, below it in every domain, and significantly below it in
24 of the 48 pairs and above it in none. In practice, however, knowing which model is stronger on one's own task also
requires setting aside part of the data with known answers and testing both models on it. Review needs this to decide
which model reviews, and using the stronger model alone needs it to decide which model to use. The selector uses the
same development questions and nothing more, so on this point the comparison is fair. Under this fair comparison, the
selector is 1.2 points [0.9, 1.4] above the stronger model alone, and 4.2 [3.8, 4.6] and 2.7 [2.4, 3.1] points above
review by the stronger and by the weaker model (Table~\ref{tab:domain-cost}).

\paragraph{Gap and headroom on held-out questions only.}
In math and knowledge, the regression of Section~\ref{sec:results} uses both development and held-out questions.
On the held-out questions alone it is unchanged: per 10 points of gap, $\Delta_S$ is lower by 1.9 [1.1, 3.2] when the
stronger model reviews and by 1.5 [0.7, 2.5] when the weaker one does; with headroom held fixed, by 1.9 [0.6, 3.5] and
1.1 [0.0, 2.3], while headroom has no detectable effect ($-0.1$ [$-3.7$, $+4.6$] and $+2.2$ [$-1.4$, $+5.7$]); and
the correlation between gap and headroom is $r = -0.73$ [$-0.84$, $-0.58$]. Since $h = (\delta - \gamma)/2$, a
larger gap leaves less headroom at a given disagreement rate $\delta$; in our pairs, $\delta$ does not grow with the
gap fast enough to offset this.

\section{Additional figures}
\label{app:figures}
\begin{figure*}[t]
  \centering
  \includegraphics[width=0.88\textwidth]{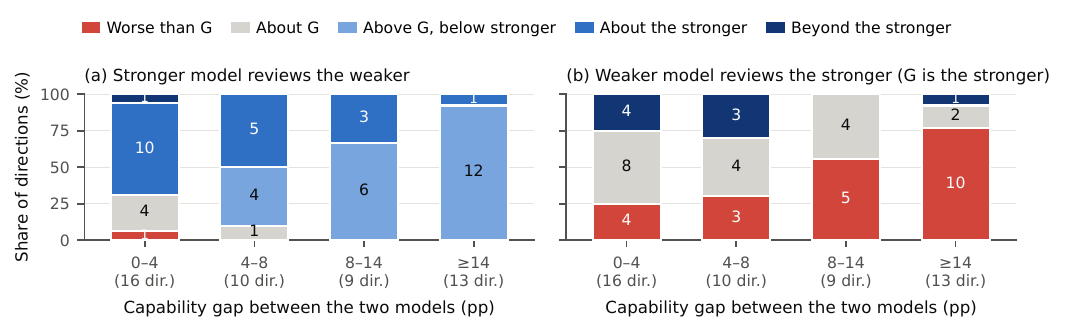}
  \caption{Where review lands, by who reviews whom and the gap. Each bar is one gap range, split by outcome level
  (paired 95\% intervals); numbers count directions. (a) The stronger model reviews the weaker. (b) The weaker reviews
  the stronger; G is then the stronger model, so only three levels occur.}
  \label{fig:levels}
\end{figure*}

\begin{figure}[t]
  \centering
  \includegraphics[width=0.72\columnwidth]{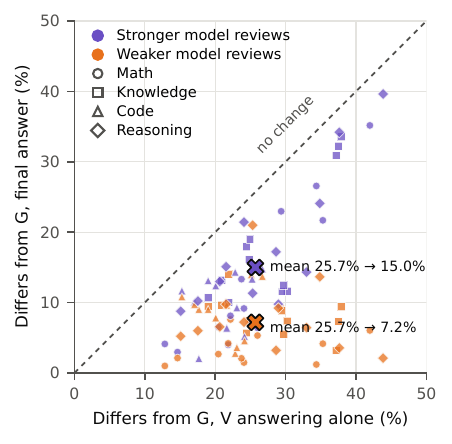}
  \caption{After review, the final answer agrees with G more often. Each point is one review direction. x: how often
  the reviewer answering alone differs from G in correctness; y: how often the final answer does. Points below the
  dashed line moved toward G; crosses are means with domains weighted equally.}
  \label{fig:convergence}
\end{figure}

\end{document}